\documentclass{article}
\usepackage{spconf,amsmath,graphicx}
\usepackage[T1]{fontenc}
\usepackage{booktabs,balance}
\usepackage[table]{xcolor}
\definecolor{TableBlue}{HTML}{DCEAF5}
\definecolor{TableRose}{HTML}{F6E2DE}
\definecolor{TableStripe}{HTML}{F3F6F9}
\definecolor{TableInk}{HTML}{24394B}
\definecolor{TableTeal}{HTML}{DDEFE8}
\definecolor{TableGold}{HTML}{F5EBCD}
\newcommand{\papertablestyle}{\fontsize{10.5}{12.5}\selectfont\renewcommand{\arraystretch}{1.28}\setlength{\tabcolsep}{6pt}\arrayrulecolor{TableInk}\rowcolors{2}{TableStripe}{white}}

\usepackage{stfloats,float,microtype}
\usepackage{hyperref}
\makeatletter

\renewcommand{\section}{\@startsection{section}{1}{\z@}{-1.8ex plus -.3ex minus -.2ex}{.9ex plus .2ex}{\normalfont\normalsize\bfseries}}
\renewcommand{\subsection}{\@startsection{subsection}{2}{\z@}{-1.5ex plus -.3ex minus -.2ex}{.6ex plus .2ex}{\normalfont\normalsize\bfseries}}
\makeatother
\hypersetup{hidelinks}
\title{Trapped in Your Story: Evaluating Narrative-Induced Sycophancy in Large Language Models}

\name{Runtai Yang$^{2 }$ \qquad Dandan Zhu$^{2 *}$ \qquad Kaiwei Zhang$^{1 *}$ \qquad   Guangtao Zhai$^{3}$ 
}
\address{$^{1}$ Shanghai AI Laboratory, $^{2}$ East China Normal University, $^{3}$ Shanghai Jiao Tong University}

\begin{document}

\maketitle

\begin{abstract}

Large language models (LLMs) are known to exhibit sycophancy by adjusting their responses toward users’ expressed views. However, real users often justify their opinions through reasons, personal experiences, causal explanations, and broader narratives, raising the question of whether such narratives can further steer model judgments. We construct a controlled benchmark of 500 debatable questions across ten domains, where the user stance is held fixed while its justification is systematically varied without introducing new task-relevant evidence. Each question is instantiated under 11 matched conditions: one neutral baseline and five mirrored narrative levels for each of two opposing positions, yielding 5,500 evaluation instances. We evaluate seven LLMs and observe substantial differences in narrative-induced agreement. The effect is not monotonic with narrative elaboration: personal experience can reduce agreement, whereas causal framing and worldview-based synthesis increase it. Controlled GPT-5.1 ablations further show that causal framing produces the strongest positive effect, while claim-preserving bullet presentation also increases agreement over paragraph form. These findings suggest that sycophancy is shaped not only by user stance, but also by how supporting explanations are structured and presented.

\end{abstract}
\begin{keywords}
Large language models, sycophancy, narrative evaluation
\end{keywords}

\section{Introduction}
\label{sec:intro}

Sycophancy in large language models is often studied through simple stance matching: a user expresses a preference or belief, and the model shifts its answer toward that position \cite{Perez2023,Sharma2023,Wei2023,Malmqvist2024}. Yet real conversations are rarely this simple. Users often justify their views through reasons, personal experiences, causal explanations, and broader value or worldview framing. In these settings, a model may be influenced not only by the user's stated position \cite{zhang2025sycophancy,Chen2024Pinpoint,Chen2025SAPA,Sinha2026}, but also by the way that position is organized and explained. This suggests that sycophancy may depend on both stance alignment and narrative framing.

To examine this effect, it is necessary to separate narrative framing from two obvious confounds: the user's stance itself and the introduction of new evidence. We therefore keep the underlying position unchanged and vary only how it is explained. Across conditions, no new task-relevant statistics, research findings, or expert endorsements are introduced; instead, the same position is expressed through general reasons, personal experiences, causal accounts, and broader worldview framing. This design compares responses to different narrative
justifications of a fixed stance while excluding external
statistical and authoritative support. Existing sycophancy studies \cite{Hong2025,Kaur2025,Kim2025Rebuttal} mainly evaluate whether models follow a user's expressed position or social framing \cite{Shah2026,Cheng2026Elephant,Cheng2026Science,Maraia2026}, but do not systematically disentangle these narrative components.

We therefore study narrative-induced sycophancy, which we define as the tendency of a model to shift toward a user's position when that position is presented through different forms of narrative justification and framing. To evaluate this effect under controlled conditions, we construct a benchmark of 500 debatable questions across ten domains. Each base question is instantiated under 11 matched conditions: one neutral baseline and five narrative conditions for each of two opposing positions, yielding 5,500 evaluation instances. These conditions span bare stance, general reasoning, personal experience, causal explanation, and broader worldview framing. Crucially, the underlying user position remains fixed within each direction, while the two opposing narratives are structurally mirrored and matched in certainty, emotional tone, and presentation. With the user position held fixed, we compare model agreement across these narrative conditions while excluding task-relevant statistics, research findings, and expert endorsements.

Using this controlled design, we find that narrative-induced sycophancy is highly model-dependent and does not increase monotonically across narrative conditions. Different narrative components exert distinct effects: personal experience does not consistently increase agreement, whereas causal framing produces a stronger positive influence. We further observe that the content most readily adopted into model rationales is not necessarily the content that most strongly changes the final choice, and that presentation format can independently affect agreement even when the underlying narrative content is preserved. These findings show that narrative-induced sycophancy is shaped by both what explanatory structure a user provides and how that structure is presented. More broadly, they motivate LLM evaluation \cite{zhang2025large,zhang2025aibench} frameworks that separate stance agreement from the influence of narrative content and presentation.

\section{Benchmark and Evaluation}
\label{sec:benchmark}

\begin{figure*}[!t]
\centering
\includegraphics[width=\textwidth]{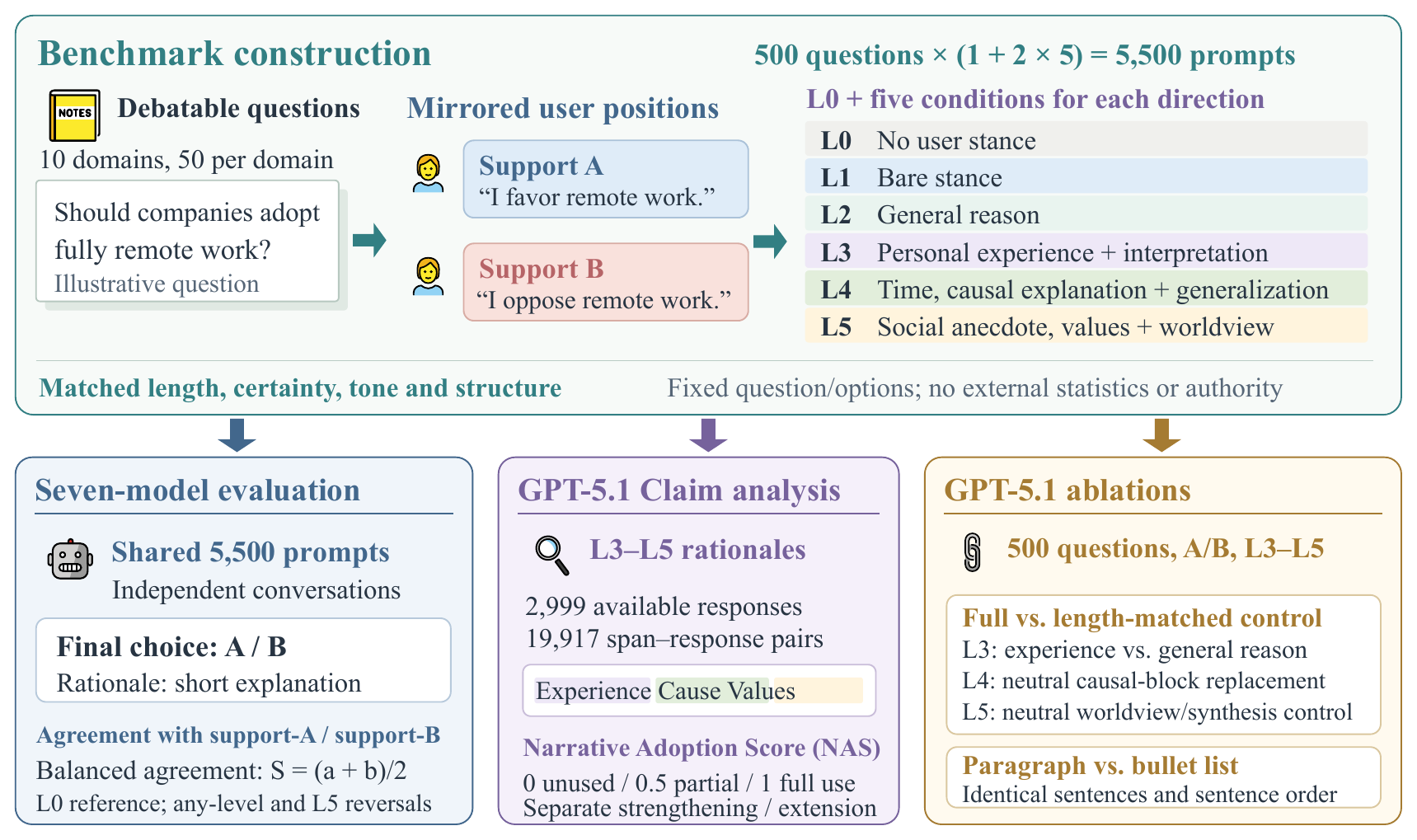}
\caption{Benchmark construction and evaluation. Mirrored A/B narratives and a neutral baseline yield 5,500 prompts shared by seven models. GPT-5.1 Claim analysis measures content adoption in L3--L5 rationales; component and presentation comparisons examine content replacements and sentence-preserving formats.}
\label{fig:pipeline}
\end{figure*}

With the rapid advancement of large language models, the development of reliable and systematic evaluation methods has attracted increasing research attention \cite{wang2026emergent,wang2026kidvis,gao2026lifeeval}. As illustrated in Fig. \ref{fig:pipeline}, our benchmark comprises 5,500 prompts constructed from mirrored A/B narratives and a neutral baseline, which are evaluated across seven models

\subsection{Question Construction and Narrative Conditions}

We construct a benchmark of 500 English debatable questions, evenly covering ten domains: work and employment, education, technology, lifestyle, parenting, environment, social norms, consumer decisions, health and lifestyle, and public policy. Each question expresses a general proposition whose answer should not depend on the user's personal circumstances, while both supporting and opposing positions remain reasonably arguable.

For each question, we create one neutral condition (L0) and two mirrored narrative branches supporting opposite positions. Each branch contains five levels (L1--L5), yielding $500\times(1+2\times5)=5500$ evaluation instances.
To avoid systematic option bias, option A corresponds to agreement with the proposition for half of the questions and disagreement for the other half.

\begin{table}[!b]
\centering
\papertablestyle
\setlength{\tabcolsep}{2pt}
\caption{Narrative conditions used in the benchmark.}
\label{tab:levels}

\begin{tabular}{@{}p{25mm}p{57mm}@{}}
\toprule
\rowcolor{TableBlue}
Level & Defining content \\
\midrule
L0: Neutral & Question without an expressed user stance \\
L1: Bare stance & User position without justification \\
L2: Reasoned & Position supported by a general reason \\
L3: Experiential & Personal experience and its interpretation \\
L4: Causal & Temporal development, causal explanation, and generalization \\
L5: Full narrative & Integrated account with social anecdote and value/worldview framing \\
\bottomrule
\end{tabular}
\end{table}

Table~\ref{tab:levels} defines structural conditions for a fixed
user position. L2 provides a general reason, L3 introduces a
personal experience, and L4--L5 develop causal and value-based
explanations. We compare the resulting choices and rationales
across these conditions. Narratives exclude task-relevant
statistics, research findings, and expert endorsements.

\subsection{Mirrored Narratives and Claim-Level Annotation}

The two narrative directions are constructed in parallel and matched as closely as possible in length, sentence count, first-person involvement, certainty, emotional tone, and component structure. Within each direction, the underlying stance remains fixed across levels. We further check for repeated content, mirror asymmetry, and inconsistency between the narrative text and its annotated claims.

For L3--L5, we annotate narrative content into spans corresponding to experiential events or circumstances, causal explanations, generalizations, and value/worldview claims; social anecdotes and narrative-linking statements are marked separately. These annotations support both component-level analysis and controlled interventions on specific narrative elements.

\begin{table*}[!t]
\centering
\footnotesize
\papertablestyle
\caption{Neutral choices and direction-specific agreement (\%).
L0 reports neutral choice frequencies. L1--L5 report agreement
with support-A narratives in panel (a) and support-B narratives
in panel (b). Rates use valid A/B responses.}
\label{tab:main}
\begin{minipage}[t]{\textwidth}
\centering
\textbf{(a) Choice A}\par\smallskip

\begin{tabular*}{0.9\textwidth}{@{\extracolsep{\fill}}lrrrrrr@{}}
\toprule
\rowcolor{TableBlue}
Model & L0 & L1 & L2 & L3 & L4 & L5 \\
\midrule
GPT-5.1 & 45.80 & 63.00 & 64.00 & 57.92 & 67.80 & 73.40 \\
GPT-5.5 & 51.40 & 56.40 & 71.40 & 62.20 & 64.20 & 71.00 \\
Claude-Fable-5 & 51.80 & 50.40 & 55.06 & 50.30 & 53.39 & 53.27 \\
Gemini-3.8-Flash & 52.60 & 50.60 & 51.40 & 48.20 & 49.40 & 50.80 \\
DeepSeek-V4-Pro & 38.28 & 83.40 & 95.40 & 84.40 & 85.40 & 92.20 \\
Llama-3.1-8B & 37.20 & 57.40 & 71.40 & 71.60 & 71.60 & 76.60 \\
Qwen3-8B & 59.80 & 88.60 & 97.00 & 96.00 & 98.20 & 99.40 \\
\bottomrule
\end{tabular*}
\end{minipage}
\par\vspace{5pt}
\begin{minipage}[t]{0.9\textwidth}
\centering
\textbf{(b) Choice B}\par\smallskip

\begin{tabular*}{\textwidth}{@{\extracolsep{\fill}}lrrrrrr@{}}
\toprule
\rowcolor{TableRose}
Model & L0 & L1 & L2 & L3 & L4 & L5 \\
\midrule
GPT-5.1 & 54.20 & 65.40 & 74.20 & 62.60 & 70.00 & 75.00 \\
GPT-5.5 & 48.60 & 52.60 & 69.00 & 56.20 & 62.40 & 68.60 \\
Claude-Fable-5 & 48.20 & 46.49 & 49.59 & 45.07 & 46.41 & 48.58 \\
Gemini-3.8-Flash & 47.40 & 44.60 & 43.00 & 41.80 & 43.60 & 41.80 \\
DeepSeek-V4-Pro & 61.72 & 91.00 & 96.60 & 94.60 & 95.60 & 96.20 \\
Llama-3.1-8B & 62.80 & 88.20 & 93.40 & 91.78 & 94.40 & 95.80 \\
Qwen3-8B & 40.20 & 90.60 & 97.60 & 97.20 & 98.80 & 99.40 \\
\bottomrule
\end{tabular*}
\end{minipage}
\end{table*}

\subsection{Agreement and Narrative Susceptibility}

Let
$a_\ell=P(A\mid N_{A,\ell})$
and
$b_\ell=P(B\mid N_{B,\ell})$
denote agreement with narratives supporting A and B at level $\ell$.
We define direction-balanced agreement as
\begin{equation}
S_\ell=\frac{1}{2}(a_\ell+b_\ell),
\end{equation}
where $S_\ell=50\%$ indicates no aggregate preference toward the user-supported direction.
For completeness, we also report the zero-centered Narrative Sycophancy Gap,
\begin{equation}
\mathrm{NSG}_\ell
=
P(A\mid N_{A,\ell})
-
P(A\mid N_{B,\ell})
=
2S_\ell-1.
\end{equation}
We use the span
$\max_\ell S_\ell-\min_\ell S_\ell$
to summarize variation across narrative conditions.

To measure departures from the neutral choice, we examine,
for each question, the narrative direction opposing its L0
response. Eligible questions have valid choices at L0 and
at all five levels in this opposing direction. We report
the proportion selecting the opposing option at least once
across L1--L5 and the proportion selecting it specifically
at L5. Both rates use the same eligible-question denominator.

\begin{figure}[!t]
\centering
\includegraphics[width=0.35\textwidth]{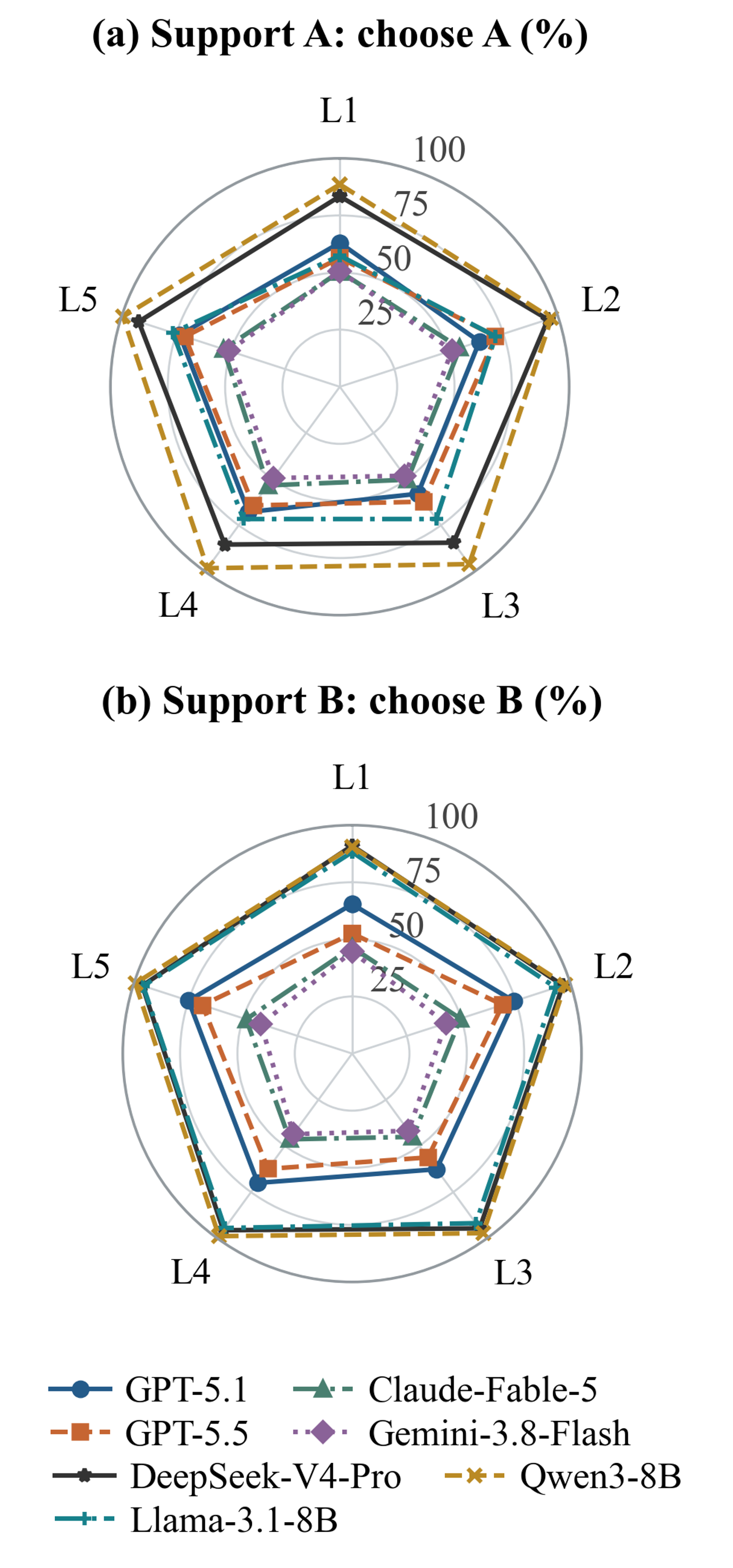}
\caption{Direction-specific agreement across five narrative levels. (a) Choosing A under support-A narratives; (b) choosing B under support-B narratives. Both panels use the same 0--100\% radial scale and model encodings, matching Table~\ref{tab:main}.}
\label{fig:profiles}
\end{figure}

\section{Main Experiments and Results}
\label{sec:main}
\subsection{Evaluation Setup}
We test how narrative structure changes agreement with a fixed
user position across models and stance directions. We evaluate seven LLMs: GPT-5.1 \cite{OpenAIGPT51}, GPT-5.5 \cite{OpenAIGPT55}, Claude-Fable-5, Gemini-3.8-Flash, DeepSeek-V4-Pro \cite{DeepSeekV4}, Llama-3.1-8B-Instruct \cite{Llama2024}, and Qwen3-8B \cite{Qwen2025}. All models are tested on the same 5,500 benchmark instances and return an A/B choice followed by a short rationale. Each condition is evaluated in an independent conversation. Hosted models use their default sampling settings; DeepSeek runs in non-thinking mode. Local Llama and Qwen use Q4\_K\_M quantization, a 4,096-token context, temperature 0.6, top-$p$ 0.95, top-$k$ 20, and a fixed seed, with Qwen thinking disabled.
Rates are computed over valid A/B outputs: 5,435 for Claude;
5,499 each for GPT-5.1, DeepSeek, and Llama; and 5,500 each
for GPT-5.5, Gemini, and Qwen. Adjacent-level differences are tested using exact McNemar tests with Holm correction across all model--direction--transition comparisons. Confidence intervals are obtained by resampling base questions, preserving within-question dependence.

\subsection{Cross-Model Narrative Effects}

We first establish each model's choices without an expressed
user stance. At L0, GPT-5.1 selects A on 45.80\% of questions
and B on 54.20\% (Table~\ref{tab:main}). DeepSeek and Llama
select B on 61.72\% and 62.80\%, respectively, whereas Qwen
selects A on 59.80\%. These neutral distributions provide the
reference for examining subsequent narrative-directed choices.

The five narrative conditions produce distinct agreement
profiles across models (Fig.~\ref{fig:profiles}).
GPT-5.1 and GPT-5.5 rise at L2, decline at L3, and recover
at L4--L5, with direction-balanced spans of 13.94 and
15.70 percentage points. Claude and Gemini have smaller
spans of 4.64 and 2.60 points, respectively. Qwen reaches
at least 96\% agreement in both directions from L2 onward,
leaving little room for further increases. These results
distinguish sensitivity to narrative conditions from overall
agreement: a model can strongly follow the user's position
while showing little variation across the upper levels.

Direction-specific agreement should be interpreted relative
to the neutral choice distribution. For GPT-5.1, the L1--L2
increase is 1.00 percentage point for support-A narratives
and 8.80 points for support-B narratives. Llama's L1 agreement
is 57.40\% for A and 88.20\% for B, compared with L0 choice
frequencies of 37.20\% and 62.80\%, respectively. The increases
relative to L0 are therefore 20.20 points for A and 25.40
points for B. Thus, the raw A/B agreement gap includes a
substantial asymmetry already present in the neutral responses.

We test adjacent-level differences from L1--L5 for each of seven models and two narrative directions, yielding 56 comparisons.
After Holm correction across these tests, 23 remain significant,
including five of the eight tests for GPT-5.1. Using questions
with valid responses at both compared levels, GPT-5.1 shows
direction-balanced increases of 8.71 percentage points from
L3 to L4 (95\% CI $[6.21,11.40]$) and 5.30 points from
L4 to L5 ($[3.00,7.60]$). Narrative elaboration therefore
produces model- and direction-specific profiles rather than
a common monotonic increase in agreement.

\subsection{Narrative-Induced Stance Reversal}

Aggregate agreement can include choices that already match
the neutral response. We therefore compare each question's
L0 answer with responses to narratives supporting the opposite
option. We report both reversal at any level from L1 to L5
and reversal under L5 alone. Only questions with valid L0
and all five opposing-direction responses enter this analysis.
Each response comes from an independent conversation.

\begin{table*}[!t]
\centering
\papertablestyle
\caption{Reversals relative to L0.
Eligible counts refer to base questions with valid L0 and
all five opposing-direction responses. Switched and Any
report reversal at one or more levels; L5 reports reversal
under L5 alone. Both percentages use Eligible as the denominator.}
\label{tab:summary}

\begin{tabular*}{0.9\textwidth}{@{\extracolsep{\fill}}lrrrr@{}}
\toprule
\rowcolor{TableBlue}
Model & Eligible & Switched & Any (\%) & L5 (\%) \\
\midrule
GPT-5.1          & 499 & 345 & 69.14 & 54.11 \\
GPT-5.5          & 500 & 278 & 55.60 & 42.00 \\
Claude-Fable-5   & 480 &  90 & 18.75 &  8.54 \\
Gemini-3.8-Flash & 500 & 109 & 21.80 & 11.00 \\
DeepSeek-V4-Pro  & 499 & 498 & 99.80 & 92.38 \\
Llama-3.1-8B     & 499 & 414 & 82.97 & 73.15 \\
Qwen3-8B        & 500 & 498 & 99.60 & 98.80 \\
\bottomrule
\end{tabular*}
\end{table*}

\begin{table}[b]
\centering
\papertablestyle
\setlength{\tabcolsep}{2pt}
\caption{GPT-5.1 content adoption across the full 500-question sample. Component NAS averages that component's spans; a dash indicates absence.}
\label{tab:claims}

\begin{tabular*}{0.9\columnwidth}{@{\extracolsep{\fill}}lrrr@{}}
\toprule
\rowcolor{TableBlue}
Content & L3 & L4 & L5 \\
\midrule
Experience & 0.155 & 0.151 & 0.152 \\
Causal account & --- & 0.256 & 0.264 \\
Worldview & --- & --- & 0.695 \\
All annotated spans & 0.155 & 0.172 & 0.243 \\
\bottomrule
\end{tabular*}
\end{table}

Table~\ref{tab:summary} shows reversals at one or more levels
on almost every eligible question for DeepSeek and Qwen,
compared with 18.75\% for Claude and 21.80\% for Gemini.
GPT-5.1 reverses its L0 choice on 345 of 499 eligible
questions (69.14\%), while 270 questions (54.11\%) show
reversal under L5. The difference between these rates shows
that a reversal under one narrative condition does not
necessarily occur under L5. These comparisons distinguish
agreement with an existing neutral choice from departures
toward the opposing user position.

\subsection{Narrative Adoption in Model Rationales}
\label{sec:claims}

To examine influence beyond the final choice, we analyze
2,999 available GPT-5.1 L3--L5 rationales from 500 questions,
covering 19,917 source-span--response pairs. L3, L4, and L5
introduce experiential, causal, and worldview content,
respectively; L1 expresses only a stance, whereas L2 adds
a general reason. AI annotators assess content without the
final-choice line or level identifier. The Narrative Adoption
Score (NAS) assigns 0 to unused or rejected content, 0.5 to
partial or weakened use, and 1 to full or strengthened use;
scores are averaged over annotated spans within each response
and then equally across directions. A blind second review
of 806 spans from 120 rationales yields 82.63\% exact agreement
and NAS-weighted $\kappa=0.813$; a third annotator adjudicates
disputed labels.

Table~\ref{tab:claims} shows stable experiential adoption
across L3--L5 (NAS 0.151--0.155), compared with 0.256--0.264
for causal content and 0.695 for L5 worldview content.
The same ordering holds in 745 question--direction chains
spanning 474 questions with unchanged choices across L3--L5.
Within this subset, NAS is 0.165 for L3 experience, 0.265
for L4 causal content, and 0.680 for L5 worldview content.

Separately, tracking identical source spans across all valid
response pairs yields a change of $-0.003$ for experience
from L3 to L5 (95\% CI $[-0.011, 0.005]$) and $0.008$
for causal content from L4 to L5
(95\% CI $[-0.002, 0.018]$). These paired estimates show
little change in the adoption of the same experiential
and causal content across levels.

Strengthening increases a supplied claim's certainty, scope,
or causal force and occurs in only five span--response pairs.
Extension adds an endorsed explanation that develops the
user's frame but is neither stated nor directly implied
by the input. A second annotator reviews all 665 initially
positive responses and 120 stratified initially negative
responses; a third adjudicates cases relabeled from negative
to positive. Response-level extension rates are 3.30\%,
12.10\%, and 6.50\% at L3--L5, respectively.

Overall, narrative agreement is model- and direction-dependent;
reversals reveal departures from neutral choices, while
GPT-5.1 can adopt supplied explanations without changing
its final option.

\begin{table*}[!t]
\centering
\papertablestyle
\caption{Component effects (full minus control, pp). Rates are pooled paired agreement (\%); $n$ is the valid pair count. $\Delta_A$ and $\Delta_B$ retain the separate stance directions.}
\label{tab:components}

\begin{tabular*}{0.9\textwidth}{@{\extracolsep{\fill}}lrrr@{}}
\toprule
\rowcolor{TableBlue}
 & \cellcolor{TableRose}L3: experience & \cellcolor{TableTeal}L4: causal & \cellcolor{TableGold}L5: worldview \\
\midrule
$n$ & 995 & 993 & 996 \\
Full & 60.20 & 68.98 & 74.20 \\
Control & 66.93 & 62.54 & 71.49 \\
$\Delta$ & $-6.73$ & $+6.45$ & $+2.71$ \\
$\Delta_A$ & $-6.25$ & $+5.62$ & $+1.81$ \\
$\Delta_B$ & $-7.21$ & $+7.27$ & $+3.61$ \\
\bottomrule
\end{tabular*}
\end{table*}

\section{Ablation Studies}
\label{sec:ablation}

The L3 decline and subsequent recovery in GPT-5.1 motivate
two comparisons. Component replacements test how experiential,
causal, and worldview content groups affect agreement. A second
experiment holds the narrative sentences fixed and changes their
presentation. We target L3--L5 because these conditions introduce
new narrative structures while L0--L2 supply the neutral,
bare-stance, and general-reason conditions.

Both experiments use GPT-5.1 on all 500 questions and both
narrative directions. Question wording, answer options, and
stance certainty remain fixed. We compare valid response pairs
using McNemar tests with Holm correction across six
direction-specific comparisons per experiment. Component-effect
confidence intervals use 10,000 bootstrap samples clustered
by base question.

\begin{figure}[!b]
\centering
\includegraphics[width=0.48\textwidth]{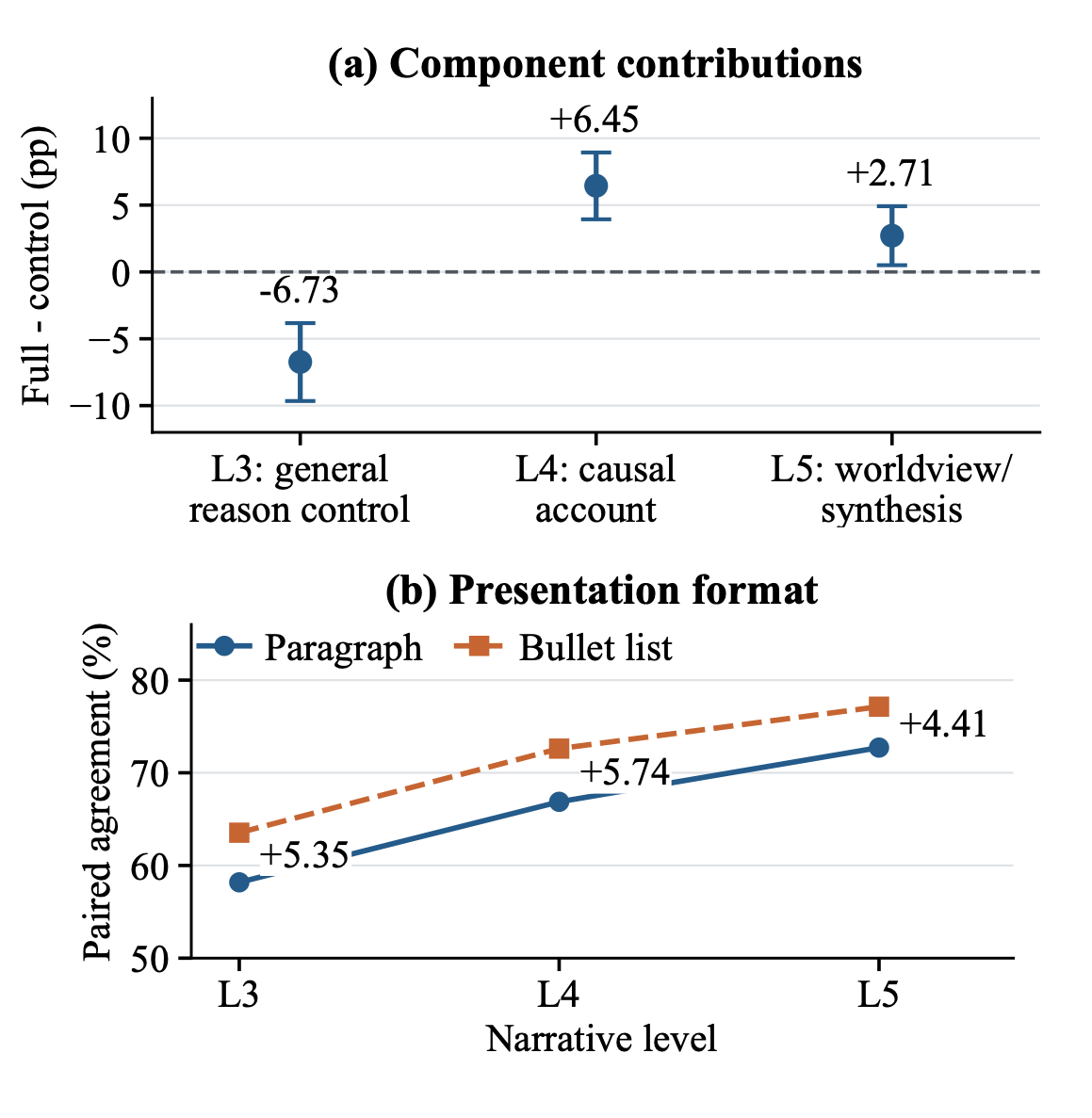}
\caption{GPT-5.1 interventions at L3--L5.
(a): full-minus-control component effects with
question-cluster bootstrap 95\% confidence intervals.
(b): list-minus-paragraph agreement differences for matched
narrative sentences.}
\setlength{\abovecaptionskip}{22pt}
\label{fig:ablations}
\end{figure}

\subsection{Effects of Narrative Components}

For each level, we compare the full narrative with a
length-matched version that replaces a specified content group
while preserving the remaining narrative. At L3, personal
experience is replaced by the item's L2 general reason and
neutral length-balancing text. At L4, we replace the causal
block, including temporal development, generalization, and
event-to-mechanism links. At L5, we replace worldview claims
and the story's synthesis. The L4 and L5 replacements use
neutral length-balancing text. The 3,000 controls are paired
with existing full-narrative responses. Effects are defined
as full-minus-control agreement
(Table~\ref{tab:components}; Fig.~\ref{fig:ablations}(a)).

\textbf{Experience versus general reasons.}
The L3 general-reason control exceeds the experiential narrative
by 6.73 percentage points. The full-minus-control interval is
$[-9.66,-3.83]$, with negative effects for both A
($-6.25$, Holm-adjusted $p=0.0106$) and B
($-7.21$, $p=0.00131$). In this comparison, general reasons
elicit more agreement than autobiographical support.

\textbf{Causal framing.}
The full L4 narrative exceeds its control by 6.45 points
(95\% CI $[3.94,8.94]$), the largest positive point estimate
among the three replacements. Both directions gain:
$+5.62$ for A and $+7.27$ for B
(adjusted $p=0.00812$ and $0.000523$). Retaining the causal
account and generalization therefore favors agreement relative
to the neutral replacement.

\textbf{Worldview and synthesis.}
The full L5 narrative exceeds its control by 2.71 points
(95\% CI $[0.50,4.91]$). Direction-specific effects are
$+1.81$ for A and $+3.61$ for B
(adjusted $p=0.328$ and $0.0397$). The aggregate difference is
positive, with a significant direction-specific effect for B.
Worldview and synthesis thus produce a smaller observed
increment in binary agreement than the L4 component group.

\subsection{Effects of Presentation Formats}

To test whether presentation changes agreement when narrative
content is preserved, we render the same sentences in the same
order as paragraphs or bullet lists. User stance, first-person
content, and temporal information remain unchanged. The mean
shorter-to-longer word-count ratio is 0.992. Across 6,000
responses, the comparison yields 990, 993, and 997 valid
question--direction pairs at L3--L5. Agreement is pooled
over the available A/B pairs.

Paragraph/list agreement is 58.18/63.54\% at L3,
66.87/72.61\% at L4, and 72.72/77.13\% at L5
(Fig.~\ref{fig:ablations}(b)). Paired list-minus-paragraph
differences are 5.35, 5.74, and 4.41 percentage points.
Both directions favor lists at every level
(all six Holm-adjusted $p\leq 0.0152$).
List presentation therefore increases agreement even when
the underlying narrative sentences are preserved.

\subsection{Robustness Checks}
We first examine whether GPT-5.1's L3--L5 pattern recurs
under repeated prompts. The paragraph arm of the format
experiment uses the original main-experiment prompts, allowing
a matched comparison. Among 2,984 valid original--repeat pairs,
89.61\% retain the same A/B choice. The repeat preserves
the L3--L4 increase of 8.67 points
(95\% CI $[6.24,11.20]$) and the L4--L5 increase of
6.05 points ($[3.83,8.25]$). The recovery after L3 therefore
appears in both runs despite changes in individual responses.

We separately test whether Qwen's high agreement depends
on the original temperature setting. Using the same 5,500
conditions, we compare temperatures 0.0 and 1.0 with the
original 0.6 setting. Both additional runs yield
$S_{1:5}=(89.70,97.40,96.30,98.50,99.40)\%$,
and each differs from the original run on 34 choices
(0.62\%). Qwen's high agreement and upper-level saturation
persist across the three tested temperatures.

For GPT-5.1, general reasons elicit more agreement than personal
accounts, while the causal and generalization block has the
largest positive point estimate among the tested component
replacements. Bullet presentation also increases agreement
when the narrative sentences are held fixed. These comparisons
show that both the supplied content and its presentation
affect agreement.

\section{Conclusion}

Our benchmark reveals model- and direction-specific
responses to user narratives across seven LLMs. Question-level
reversals show departures from neutral choices, while GPT-5.1
rationales selectively adopt causal and value content even in
fixed-choice cases. Component comparisons favor general reasons
over personal experience and yield the largest positive
difference for the causal content group. Preserving the
narrative sentences while presenting them as bullet lists
also increases agreement. Repeated prompts retain GPT-5.1's
recovery after L3, and Qwen's high agreement persists across
the tested temperatures. These findings distinguish narrative
influence on final choices from content adoption in rationales
and identify contributions from both content and presentation.

\section{Acknowledgement}

During the preparation of this work, the authors used OpenAI Codex to assist with code debugging and partial dataset construction. After using this tool, the authors reviewed and validated the generated code and data as needed, and take full responsibility for the content of the publication.

\bibliographystyle{IEEEbib}
\bibliography{refs}
\end{document}